\documentclass[runningheads]{llncs}

\usepackage{eccv}
\usepackage{wrapfig}
\usepackage{tabularx}
\usepackage{array}

\usepackage{eccvabbrv}

\usepackage{graphicx}
\usepackage{booktabs}

\usepackage[accsupp]{axessibility}  

\usepackage{booktabs}
\usepackage[table]{xcolor}
\usepackage{multirow}

\definecolor{spotegray}{gray}{0.90}
\definecolor{gainGreen}{RGB}{0,128,0}

\newcommand{\gain}[1]{\textcolor{gainGreen}{\scriptsize\,#1$\uparrow$}}

\usepackage{hyperref}

\usepackage{orcidlink}

\begin{document}

\title{SPOT-E: Test-Time Entropy Shaping with Visual Spotlights for Frozen VLMs} 

\titlerunning{Test-Time Entropy Shaping with Visual Spotlights for Frozen VLMs}

\author{Bo Yin\inst{1} \and Xiaobin Hu\inst{1}\and Chengming Xu\inst{2}\and Ruolin Shen\inst{3}\and Mo Yang\inst{4}\and Jiangning Zhang\inst{5}\and Peng-Tao Jiang\inst{6}\and Cheng Tan\inst{7}\and Shuicheng Yan\inst{1}}

\authorrunning{Yin et al.}

\institute{National University of Singapore \and Fudan University \and Technical University of Munich \and Sagenic Tech \and Zhejiang University \and vivo \and Shanghai Artificial Intelligence Laboratory}

\maketitle

\begin{abstract}
Vision-language models (VLMs) often underperform on evidence intensive tasks because decisive visual evidence are small, localized, and easy to overlook, leading to failures in evidence readout even when high-level reasoning is intact. 
Prior inference-time visual interventions can improve grounding without retraining, but they are largely open-loop and lack a mechanism to verify whether highlighted evidence is actually used. 
We study answer-span prediction entropy as a model-internal feedback signal and show that naive entropy minimization is ambiguous, since low entropy may arise from evidence-grounded confidence or shortcut collapse. 
To resolve this ambiguity, we introduce low-entropy anchors and an entropy-shaping objective that reduces answer uncertainty while preserving baseline high-confidence tokens. 
We instantiate this principle in SPOT-E, a plug-and-play test-time method that produces question-conditioned spotlights, optimized per instance via light-weight tuning based on Group Relative Policy Optimization (GRPO). 
Across all benchmarks and different VLM families, SPOT-E yields consistent gains and improved robustness under visual corruptions.
Code is publicly available at: \url{https://github.com/YinBo0927/SPOT-E}
\end{abstract}

\section{Introduction}
Vision-language models have made rapid progress in multimodal understanding, yet they remain unreliable on evidence-intensive tasks such as chart reading and document parsing~\cite{liu2023visual, masry2022chartqa, mathew2021docvqa, li2026vision, yin2026policy, yu2026latent, hu2026landscape}.
In these settings, the decisive evidence are often small and localized.
A model may describe a correct reasoning plan, for example it may say ``read the y-axis value and then compare the two bars'', but still misread the underlying number, causing the final answer to fail~\cite{lu2023mathvista, zhang2023multimodal, guan2024hallusionbench, xing2026boosting, yu2026vismem, liu2025human, yu2025visual, yu2026visual}.
This pattern exposes a bottleneck that we call the evidence utilization gap: the model can reason about what evidence is needed, but cannot reliably extract and focus on the fine-grained visual evidence that determines the answer~\cite{tong2024eyes}.
Closing this gap for frozen, already-deployed VLMs, without costly retraining or task-specific annotation, is practically urgent and methodologically challenging~\cite{yin2024woodpecker}.
\textbf{How can we improve evidence utilization at inference time while keeping the backbone frozen?}

A natural direction is to intervene on the visual input at test time so that decisive regions become more accessible to the model~\cite{yang2023set, zhang2023visual, yin2025fera, snell2024scaling, brown2024large}.
Methods such as FGVP instantiate this idea and report gains without modifying model weights~\cite{yang2023fine, yu2026dual}.
However, these approaches are open-loop.
They apply a fixed intervention but do not provide a mechanism to verify whether the model actually relied on the emphasized evidence.
When the selected region misses the decisive evidence, or when the intervention degrades the evidence itself, the failure remains invisible to the method and therefore cannot be corrected.
This raises a key question: \textbf{how can we design a test-time intervention that is both instance-adaptive and self-verifying?}

These limitations motivate a closed-loop feedback signal that can be obtained from a frozen VLM during inference~\cite{shinn2023reflexion, jian2025look}.
We find that answer entropy, computed from the logits over the tokens that form a structured final answer, tracks evidence usability.
Entropy is low when the decisive visual cues are clear, and it increases when they are obscured~\cite{kadavath2022language, kuhn2023semantic, xiong2023can, yin2026refinement}.
However, entropy reduction is inherently ambiguous.
Both evidence-grounded confidence and shortcut behaviors can produce low entropy~\cite{geirhos2020shortcut,carter2021overinterpretation, yin2025don}, as shown in Fig.~\ref{fig:entropy_ambiguity}.
As a result, \textbf{minimizing entropy alone can drive the model into confident-but-wrong shortcuts}, creating a systematic failure mode in the absence of labels.

To resolve this ambiguity, we introduce an \textbf{entropy-shaping} principle that reduces answer entropy while preserving the model's baseline high-confidence predictions on the unmodified input.
Specifically, we identify low-entropy anchors as token positions where the frozen VLM is already near-deterministic at baseline, and we penalize interventions that disrupt these anchors.
This yields a label-free objective that favors evidence-supported confidence over shortcut-induced collapse.
We instantiate the principle in \textbf{SPOT-E}, a test-time visual adaptation framework that augments a frozen VLM with a lightweight, question-conditioned visual spotlight module that produces spotlights.
For each instance, we optimize only the spotlight-module LoRA~\cite{hu2022lora} parameters using Group Relative Policy Optimization~\cite{shao2024deepseekmath} under the entropy-shaping reward.

Our contributions are summarized as follows:
\begin{itemize}
\item \textbf{Entropy Signal.} We identify answer entropy as a label-free signal for evidence utilization in frozen VLMs, and show that entropy reduction is inherently ambiguous.

\item \textbf{Entropy Shaping.} We propose low-entropy anchors and an entropy-shaping reward to disambiguate evidence-supported confidence from shortcut collapse.

\item \textbf{SPOT-E Framework.} We present SPOT-E, a plug-and-play test-time framework that keeps the VLM frozen while optimizing a question-conditioned visual spotlight via per-instance GRPO to leverage low-entropy anchors.

\item \textbf{Broad Evaluation.} We conduct extensive evaluations across both open-source and closed-source VLM families and multiple backbones, demonstrating consistent gains on diverse benchmarks, with particularly strong improvements on evidence-intensive tasks and improved robustness under corruptions.
\end{itemize}
\section{Related Work}

\subsection{Inference-Time Visual Interventions and Visual Prompting}
Inference-time visual interventions improve grounding by manipulating the visual evidence presented to a frozen VLM at test time~\cite{bar2022visual, chen2023understanding}.
Common approaches include overlaying marks or masks to highlight regions, as in FGVP and Set-of-Mark prompting, selectively cropping or zooming into candidate areas to preserve small details, as in ViCrop, and spatial transformations that reallocate resolution toward query-relevant evidence, as in AttWarp~\cite{yang2023fine, yang2023set, zhang2023visual, dalal2025constructive}.
Another practical direction uses attention- or score-guided prompting strategies such as API-style prompting to steer the model toward informative regions without retraining~\cite{yu2024attention}.
These methods are appealing because they are lightweight and model-agnostic, but they often depend on fixed heuristics, external region proposals, or discrete choices that may be brittle across instances.
Our method fits this paradigm and uses a question-conditioned spotlight for adaptive evidence emphasis under a frozen backbone.

\subsection{Entropy and Uncertainty for Evidence Localization}
Uncertainty signals have long been used to diagnose and steer model behavior at inference time~\cite{kadavath2022language, hendrycks2016baseline}.
Entropy over output distributions is a common proxy for confidence and has been used for calibration and selective prediction, self-consistency and re-ranking, as well as entropy- and confidence-driven decoding heuristics~\cite{guo2017calibration, wang2022self, kuhn2023semantic}.
In vision and multimodal reasoning, uncertainty is also closely tied to evidence localization, since failures in fine-grained grounding often appear as high uncertainty concentrated on a small set of answer tokens~\cite{li2023evaluating, huang2024opera}.
Recent multimodal work further leverages entropy or confidence to trigger additional perception, guide region selection, and filter visually unsupported generations~\cite{yin2024woodpecker, yin2023large}.
Our method combines an entropy-shaping reward with a question-conditioned visual spotlight to encourage decisive, visually supported answers without updating the frozen VLM.
\section{Motivation}
\label{sec:motivation}
In this section, we motivate an entropy-centric view of visual adaptation in VLMs.
When decisive visual evidence is usable, the model commits more consistently to a single final answer, making answer-span entropy a proxy for evidence use.
We show that this effect is spatially grounded and can change sharply with the visibility of the decisive region.
However, entropy reduction is ambiguous and may also reflect shortcut behaviors that suppress hard evidence, yielding confident-but-wrong predictions.
To resolve this, we introduce low-entropy anchors and an entropy-shaping principle that reduces answer entropy while preserving anchor stability.

\subsection{Notation}
\label{sec:background}

Consider a frozen VLM $F_\phi$, which typically comprises a visual encoder $V$, a multimodal connector $F_c$, and an LLM $M$.
Given an input image $x$ and a user instruction $q$, the model receives a multimodal sequence of visual tokens $\{v_1,v_2,\ldots,v_n\}$ and text tokens $\{t_1,t_2,\ldots,t_m\}$.
During decoding, $F_\phi$ processes the concatenated multimodal context
$\{v_1,\ldots,v_n,t_1,\ldots,t_m\}$ followed by previously generated tokens $\{y_1,\ldots,y_{k-1}\}$ to predict the next token $y_k$.

\noindent
\textbf{Next-token distribution.}
At decoding step $k$, the model outputs a distribution over the vocabulary $\mathcal{W}$:
\begin{equation}
p_k(w) = p_\phi\!\left(w \mid x,q,y_{<k}\right), \qquad w\in\mathcal{W},
\label{eq:next_token_dist}
\end{equation}
where $y_{<k}=(y_1,\ldots,y_{k-1})$ is the prefix.

\noindent
\textbf{Entropy.}
We quantify predictive uncertainty using Shannon entropy of the next-token distribution:
\begin{equation}
H_k(x,q) = -\sum_{w\in\mathcal{W}} p_k(w)\log p_k(w).
\label{eq:token_entropy}
\end{equation}
Unless otherwise specified, $\log(\cdot)$ denotes the natural logarithm.
We compute entropy under the baseline decoding trajectory.

\noindent
\textbf{Answer entropy.}
Long-form generations contain many tokens weakly related to visual evidence.
To focus on evidence-relevant uncertainty, we compute entropy on a final answer span.
We enforce a structured output format (e.g., \texttt{Final answer:} \emph{...}) and extract the token indices of the answer span as $\mathcal{T}_{\mathrm{ans}}$.
We define answer entropy as
\begin{equation}
H_{\mathrm{ans}}(x,q) = \frac{1}{|\mathcal{T}_{\mathrm{ans}}|}\sum_{k\in\mathcal{T}_{\mathrm{ans}}} H_k(x,q).
\label{eq:answer_entropy}
\end{equation}
For an intervention producing a modified input $\tilde{x}$, we measure entropy reduction by
\begin{equation}
\Delta H_{\mathrm{ans}} = H_{\mathrm{ans}}(\tilde{x},q)- H_{\mathrm{ans}}(x,q).
\label{eq:delta_answer_entropy}
\end{equation}

\noindent
\textbf{Low-entropy anchors.}
To characterize token positions that the base model is already confident about, we define a set of low-entropy anchors under the baseline input.
Concretely, we select the $K$ positions with the smallest next-token entropies:
\begin{equation}
\mathcal{I}_{\mathrm{low}}(x,q)= 
\mathrm{TopK_{small}}\big(\{H_k(x,q)\}_{k=1}^{T}\big),
\label{eq:low_entropy_anchor}
\end{equation}
where $T$ is the output length. In addition, we evaluate $H_{k}(\tilde{x},q)$ by conditioning on the baseline token prefix to align anchor positions.
These anchors are later used to distinguish desirable entropy reduction from shortcut behaviors.

\subsection{Visual Evidence Shapes Answer Entropy}
A natural intuition is that visual evidence affects a VLM's answer mainly by changing how certain the model can be at the point of committing to the final answer.
If the decisive evidence for $q$ is clear, the model should concentrate probability mass on a consistent answer; if the evidence is weak or obscured, the model should remain uncertain.
We operationalize this intuition using the answer entropy $H_{\mathrm{ans}}(x,q)$ in Eq.~\eqref{eq:answer_entropy}, which measures uncertainty on the final answer.

\begin{figure*}[t]
    \centering
    \includegraphics[width=0.85\textwidth]{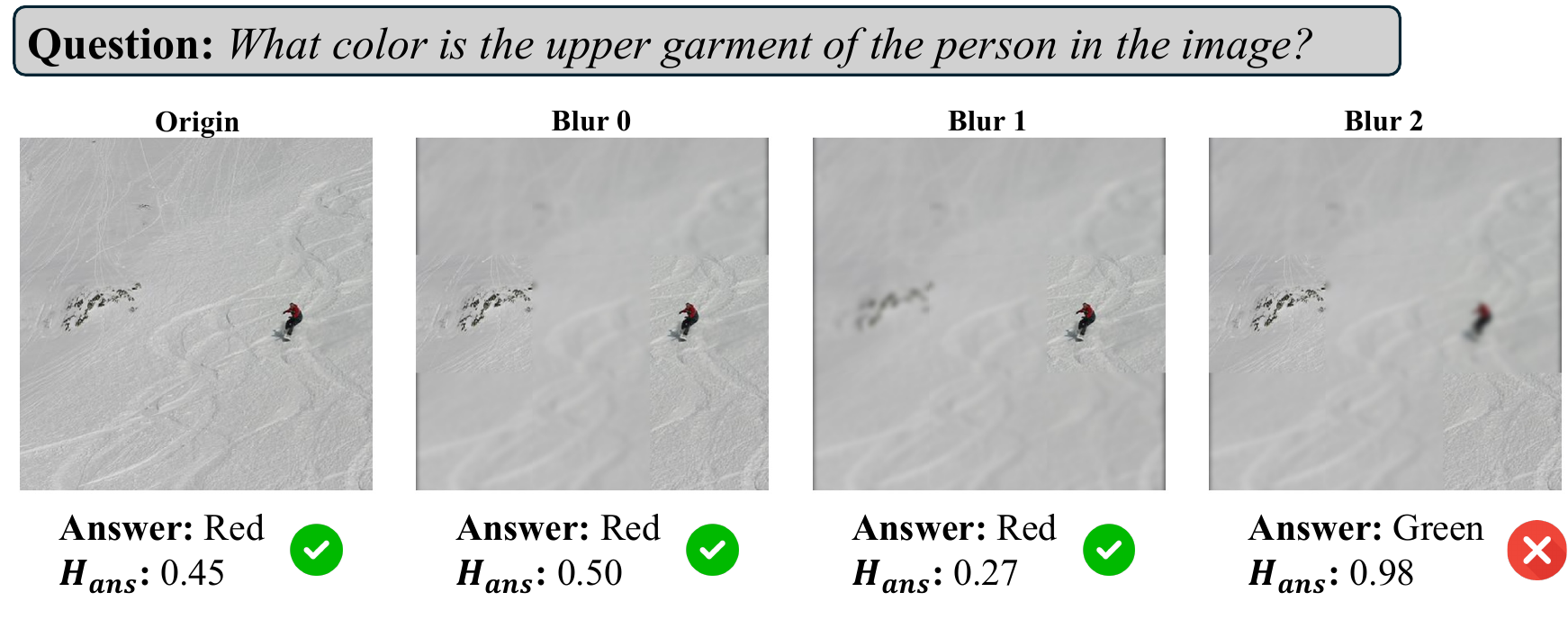}
    \caption{Localized evidence controls answer entropy.
    We apply region-level blur to a subset of grid regions while keeping all other pixels unchanged, and measure $H_{\mathrm{ans}}(\tilde{x}_S,q)$.}
    \label{fig:entropy_sensitivity_example}
\end{figure*}

We validate the intuition with a simple spatial sensitivity analysis.
We partition the image into a coarse grid of regions $\mathcal{R} = \{r_1, \dots, r_M\}$ and apply localized interventions while keeping the rest of the input unchanged.
For a given subset size $n$, we sample subsets $S \subseteq \mathcal{R}$ with $|S| = n$, construct an intervened image $\tilde{x}_S$ by applying a fixed region-level transformation only within regions in $S$ such as blurring, and compute $H_{\mathrm{ans}}(\tilde{x}_S, q)$.
Fig.~\ref{fig:entropy_sensitivity_example} shows a representative instance where suppressing the localized evidence required by $q$ increases $H_{\mathrm{ans}}$, eventually leading to incorrect predictions.

\subsection{Entropy Reduction is Ambiguous}
The previous subsection suggests that lowering $H_{\mathrm{ans}}$ often correlates with making decisive evidence more usable.
However, entropy reduction alone is ambiguous: an intervention may decrease uncertainty not by improving evidence quality, but by attenuating hard evidence and steering the model toward priors or salient distractors, yielding confident-but-wrong answers with smaller $H_{\mathrm{ans}}$.

We illustrate this ambiguity within the same subset-based intervention framework, but now restricting attention to the most entropy-sensitive singleton subset.
Concretely, for any subset $S \subseteq \mathcal{R}$, we define the entropy change induced by intervening on $S$ as
\begin{equation}
\Delta H_{\mathrm{ans}}(S)
\;=\;
 H_{\mathrm{ans}}(x, q) - H_{\mathrm{ans}}(\tilde{x}_S, q) ,
\label{eq:delta_h_ans_def}
\end{equation}
where $\tilde{x}_S$ modifies only regions in $S$ and leaves all other pixels unchanged. We identify
\begin{equation}
S^\star
=
\arg\max_{|S|=1}
\left|
\Delta H_{\mathrm{ans}}(S)
\right|,
\label{eq:s_star_def}
\end{equation}
i.e., the single-region subset whose intervention yields the largest magnitude of entropy change.
We then keep the rest of the image fixed and construct a family of intervened inputs by blurring only within $S^\star$ with increasing strength $\alpha$:
\begin{equation}
\tilde{x}_{S^\star}^{(\alpha)} = \mathcal{A}_\alpha(x;\,S^\star),
\qquad \alpha \in [0,1],
\label{eq:evidence_blur_sweep}
\end{equation}
where $\mathcal{A}_\alpha(\cdot;\,S^\star)$ applies blur of level $\alpha$ to regions in $S^\star$ and leaves all other regions unchanged.

\begin{figure}[t]
    \centering
    \includegraphics[width=\textwidth]{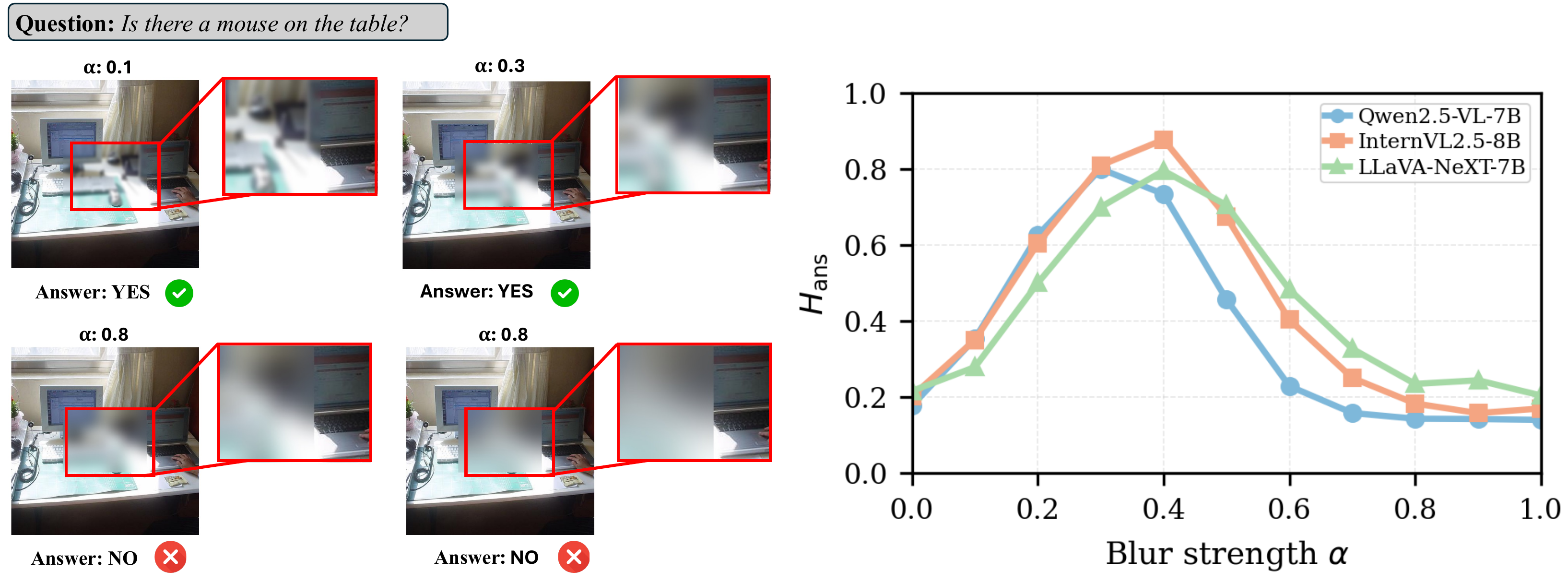}
    \caption{\textbf{Entropy reduction can be misleading.}
    We blur only the most entropy-sensitive singleton subset $S^\star$ with strength $\alpha$.
    $H_{\mathrm{ans}}$ often rises as evidence becomes ambiguous, but may drop again when evidence is erased.}
    \label{fig:entropy_ambiguity}
\end{figure}

Sweeping $\alpha$ typically yields a non-monotonic profile:
$H_{\mathrm{ans}}$ is low when the evidence is clear (small $\alpha$), peaks when the evidence becomes ambiguous (intermediate $\alpha$), and can decrease again once the evidence is effectively erased (large $\alpha$), as the model collapses to a prior- or distractor-driven answer.
Fig.~\ref{fig:entropy_ambiguity} shows that low $H_{\mathrm{ans}}$ at large $\alpha$ can also arise from prior-driven overconfidence.

\subsection{Entropy Shaping with Low-Entropy Anchors}
\label{sec:obs3}
\begin{figure}[t]
  \centering
  \includegraphics[width=\linewidth]{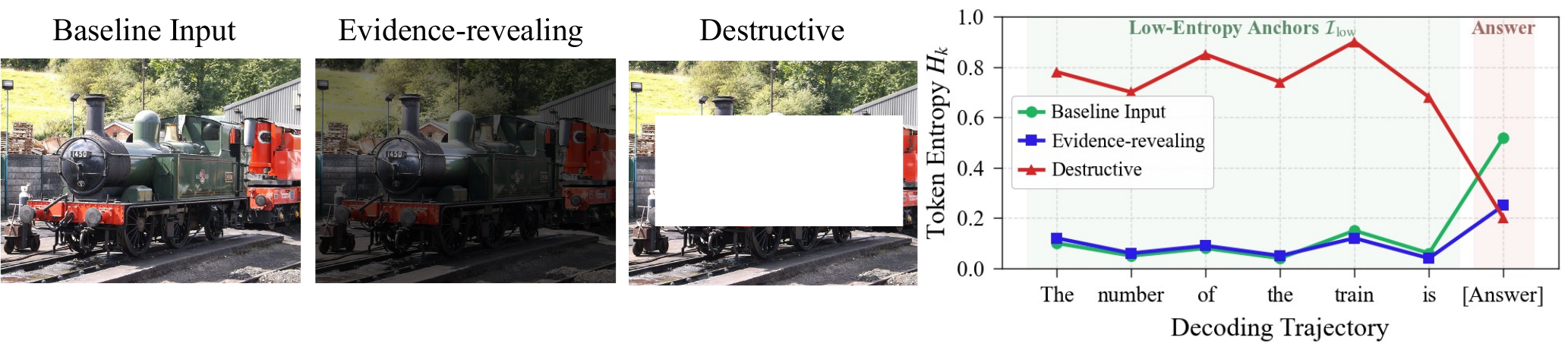}
  \caption{Low-entropy anchors reveal destructive shortcuts.}
  \label{fig:low_entropy_anchors}
\end{figure}

Answer entropy $H_{\mathrm{ans}}$ provides a convenient scalar readout of how concentrated the model is on the final answer.
The difficulty is that the same decrease in $H_{\mathrm{ans}}$ can be produced by different mechanisms: the intervention may genuinely expose missing evidence, or it may remove hard evidence and let the model settle on priors or distractors.
To disambiguate these cases, we look beyond the answer entropy itself and ask whether an intervention is non-destructive, namely whether it preserves parts of the decoding process that the base model was already confident about. As illustrated in Fig.~\ref{fig:low_entropy_anchors}, two interventions may achieve a similar drop in answer entropy, yet only the evidence-revealing one preserves the baseline's low-entropy tokens, whereas the destructive shortcut inflates their entropy motivating our anchor disruption measure.

We use low-entropy anchors $\mathcal{I}_{\mathrm{low}}(x,q)$ in Eq.~\eqref{eq:low_entropy_anchor} to represent such stable positions under the baseline input.
Given an intervened input $\tilde{x}$, we measure anchor disruption by the average entropy increase on anchor positions,
\begin{equation}
\Delta H_{\mathrm{low}}(\tilde{x}) = \frac{1}{|\mathcal{I}_{\mathrm{low}}|}\sum_{k\in \mathcal{I}_{\mathrm{low}}}\max\!\big(0,\; H_k(\tilde{x},q)-H_k(x,q)\big).
\label{eq:delta_entropy_low}
\end{equation}
Interventions with comparable reductions in $H_{\mathrm{ans}}$ can behave very differently under this criterion.
Evidence-revealing interventions tend to keep $\Delta H_{\mathrm{low}}$ small, while shortcut interventions often reduce $H_{\mathrm{ans}}$ at the cost of increasing entropy on anchors. This motivates an entropy-shaping principle that favors reducing $H_{\mathrm{ans}}$ while preserving low-entropy anchors.

\section{SPOT-E: Visual Spotlighting for Entropy Shaping}
\begin{figure}[t]
    \centering
    \includegraphics[width=0.95\textwidth]{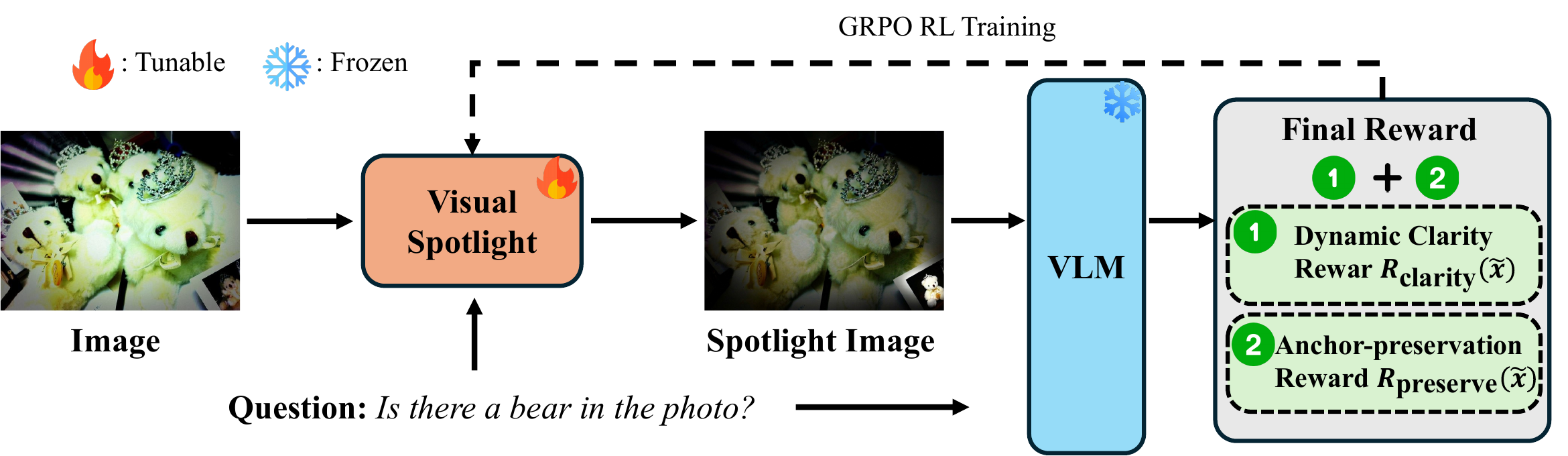}
    \caption{\textbf{SPOT-E overview.} SPOT-E freezes the VLM and optimizes a lightweight visual spotlight at test time to generate an intervened image, scored by answer-entropy clarity and anchor-preservation.}
    \label{fig:overview}
\end{figure}

Building on Sec.~\ref{sec:motivation}, we introduce \textbf{SPOT-E}.
As shown in Fig.~\ref{fig:overview}, SPOT-E performs entropy-guided test-time visual adaptation by keeping the VLM $F_\phi$ frozen and optimizing a lightweight, question-conditioned visual spotlight.
Given an image $x$ and instruction $q$, the spotlight produces an intervened image $\tilde{x}=\mathcal{S}(x;,m)$, where $\mathcal{S}$ is the spotlighting operator and $m$ is a soft pixel mask.
We feed $\tilde{x}$ into $F_\phi$ to compute the answer entropy $H_{\mathrm{ans}}(\tilde{x},q)$ and anchor entropies on $\mathcal{I}_{\mathrm{low}}(x,q)$.
At test time, SPOT-E runs a short per-instance episode: it samples candidate spotlights, scores them with an entropy-shaping objective, and updates only the spotlight parameters via GRPO~\cite{shao2024deepseekmath}.
The final prediction is chosen by Best-of-$N$ over candidates, and the spotlight is reset after each instance.

\subsection{Visual Spotlight}
\label{sec:spotlight}

\begin{figure}[t]
    \centering
    \includegraphics[width=0.95\textwidth]{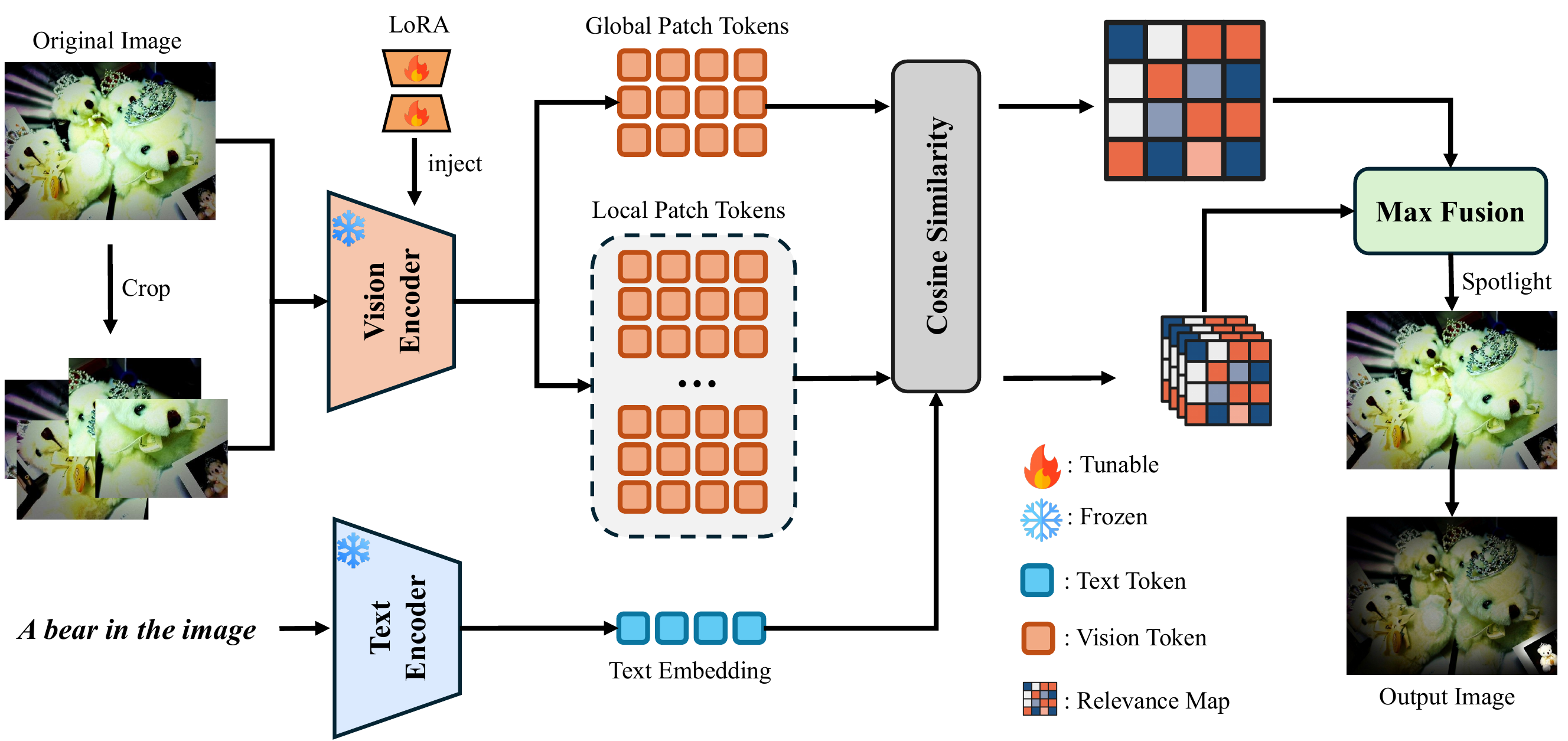}
    \caption{\textbf{SPOT-E visual spotlight module.} Both the image encoder and the text encoder are CLIP. The module computes patch-text similarities on the global view and local crops, then fuses multi-view relevance maps via max pooling to produce the final spotlight mask.}
    \label{fig:spote_eye}
\end{figure}

As illustrated in Fig.~\ref{fig:spote_eye}, SPOT-E introduces a CLIP-based~\cite{radford2021learning} visual spolight module to produce question-conditioned visual spotlights.
Given instruction $q$, we extract a compact visual phrase $\bar{q}$ by retaining the key entities and attributes relevant to visual grounding. Then the module extracts global patch tokens and local crop tokens with a CLIP vision encoder with LoRA~\cite{hu2022lora} adapter, matches them to the frozen CLIP text embedding via patch–text similarity to obtain relevance maps, and fuses multi-view evidence by max pooling to form the final spotlight mask.

\noindent\textbf{Global view and crop max fusion.}
We run the CLIP on the full image and on multiple crops to avoid missing small evidence.
Let $x^{(0)}= x$ be the global view and $\{x^{(i)}\}_{i=1}^{N_c}$ be crops.
For each view $x^{(i)}$, CLIP outputs patch tokens $\{p^{(i)}_j\}_{j=1}^{N_i}$ and a text embedding $t(\bar{q})$.
We compute patch relevance by cosine similarity:
\begin{equation}
u^{(i)}_j
\;=\;
\Big\langle \mathrm{norm}\!\big(p^{(i)}_j\big),\; \mathrm{norm}\!\big(t(\bar{q})\big)\Big\rangle.
\label{eq:clip_sim_map}
\end{equation}
We reshape $\{u^{(i)}_j\}$ to a 2D grid and map each crop grid back to the full-image coordinates via $\mathcal{W}_i(\cdot)$, which warps the $i$-th crop map to the global image coordinates.
The fused relevance map is obtained by max fusion:
\begin{equation}
u \;=\; \max\!\Big( u^{(0)},\; \max_{i=1,\ldots,N_c}\mathcal{W}_i\!\left(u^{(i)}\right) \Big).
\label{eq:max_fusion}
\end{equation}

\noindent\textbf{Soft mask and spotlighting operator.}
We upsample the fused relevance map $u$ to the image resolution and obtain a soft pixel mask
\begin{equation}
m \;=\; \sigma\!\left(\tfrac{1}{\tau}\,u^{\uparrow}\right)\in[0,1]^{H\times W},
\label{eq:spotlight_mask}
\end{equation}
where $u^{\uparrow}$ denotes bilinear upsampling of $u$ and $\tau$ controls mask sharpness.
We then form the intervened input via
\begin{equation} \tilde{x} \;=\; \mathcal{S}(x;\,m) \;=\; m\odot x \;+\; (1-m)\odot \mathcal{B}(x), \label{eq:spotlight_operator} \end{equation} where $\odot$ denotes element-wise multiplication and $\mathcal{B}(\cdot)$ is a fixed background-degrading transform (background dimming).

\subsection{Entropy-Shaping Reward}
\label{sec:reward}

Given a candidate spotlight mask $m$ and intervened input $\tilde{x}=\mathcal{S}(x;\,m)$, we score it using an entropy-shaping reward
\begin{equation}
R(\tilde{x}) \;=\; R_{\mathrm{clarity}}(\tilde{x}) \;+\; R_{\mathrm{preserve}}(\tilde{x}).
\label{eq:reward_total}
\end{equation}
Both terms are computed from the frozen VLM logits, while gradients update only the visual spotlight module.

\noindent\textbf{Dynamic clarity reward.}
We encourage the model to become more decisive on the final answer span by reducing answer entropy:
\begin{equation}
\Delta H_{\mathrm{ans}}(\tilde{x}) \;=\; H_{\mathrm{ans}}(x,q)-H_{\mathrm{ans}}(\tilde{x},q).
\label{eq:delta_ans_method}
\end{equation}
To avoid over-optimizing when the base model is already confident, we apply a dynamic scaling factor based on the baseline answer entropy:
\begin{equation}
R_{\mathrm{clarity}}(\tilde{x})
\;=\;
\gamma(x,q)\cdot \Delta H_{\mathrm{ans}}(\tilde{x}),
\qquad
\gamma(x,q)\;=\;\frac{H_{\mathrm{ans}}(x,q)}{H_{\mathrm{ans}}(x,q)+c},
\label{eq:reward_clarity}
\end{equation}
where $c>0$ is a small constant. When the baseline is uncertain, $\gamma$ increases and entropy reduction is rewarded more; when the baseline is already confident, $\gamma$ suppresses blind exploration.

\noindent\textbf{Anchor-preservation reward.}
Entropy reduction can be achieved by shortcut behaviors that disrupt tokens the base model was already confident about.
We therefore penalize entropy increases on low-entropy anchor positions $\mathcal{I}_{\mathrm{low}}(x,q)$ (Eq.~\eqref{eq:low_entropy_anchor}) using the anchor disruption measure $\Delta H_{\mathrm{low}}(\tilde{x})$ (Eq.~\eqref{eq:delta_entropy_low}):
\begin{equation}
R_{\mathrm{preserve}}(\tilde{x})
\;=\;
-\lambda\cdot \Delta H_{\mathrm{low}}(\tilde{x}),
\label{eq:reward_preserve}
\end{equation}
where $\lambda$ controls the strength of anchor preservation.


\subsection{Test-Time Optimization with GRPO}
\label{sec:grpo}

SPOT-E runs a short per-instance test-time optimization episode to adapt the visual spotlight module, while keeping the VLM $F_\phi$ fully frozen.
Let $\theta$ denote the visual spotlight parameters (LoRA adapters in CLIP vision encoder attention layers), initialized to $\theta_0$ for each instance.

\noindent\textbf{Group sampling and scoring.}
At each iteration, we sample a group of $N$ candidate masks by injecting Gaussian noise into the visual spotlight and obtain intervened inputs $\{\tilde{x}^{(n)}\}_{n=1}^N$.
Each candidate is scored by the total reward $R(\tilde{x}^{(n)})$ (Eq.~\eqref{eq:reward_total}).

\noindent\textbf{Group-relative advantages.}
We compute standardized advantages within the group:
{\footnotesize
\begin{equation}
\mu_R = \frac{1}{N}\sum_{n=1}^{N} R(\tilde{x}^{(n)}), \qquad
\sigma_R = \sqrt{\frac{1}{N}\sum_{n=1}^{N}\big(R(\tilde{x}^{(n)})-\mu_R\big)^2}.
\label{eq:grpo_stats}
\end{equation}
\begin{equation}
A^{(n)} = \frac{R(\tilde{x}^{(n)})-\mu_R}{\sigma_R+\epsilon}.
\label{eq:grpo_adv}
\end{equation}
}
where $\epsilon$ is a small constant.

\noindent\textbf{GRPO update and reset.}
We apply a standard GRPO clipped policy update on $\theta$ using $\{A^{(n)}\}$, with a KL regularizer to keep the visual spotlight close to its initialization.
{\footnotesize
\begin{equation}
\bar{r}^{(n)}(\theta)=\mathrm{clip}\!\big(r^{(n)}(\theta),\,1-\delta,\,1+\delta\big).
\label{eq:grpo_clip}
\end{equation}
\begin{equation}
\mathcal{L}_{\mathrm{GRPO}}(\theta) = -\frac{1}{N}\sum_{n=1}^{N} \min\!\Big(r^{(n)}(\theta)A^{(n)},\,\bar{r}^{(n)}(\theta)A^{(n)}\Big)+\beta\,\mathrm{KL}\!\big(\pi_\theta\,\|\,\pi_{\theta_0}\big).
\label{eq:grpo_obj}
\end{equation}
}
where $\delta$ is the clipping threshold and $\beta$ controls the KL strength.
We update only $\theta$ by gradient descent on Eq.~\eqref{eq:grpo_obj}, and reset $\theta\leftarrow\theta_0$ after each instance to avoid cross-sample drift.
\section{Experiments}
\label{sec:experiments}
\noindent\textbf{Overview.}
We evaluate SPOT-E from four complementary angles.
First, we report main results across a broad set of frozen backbones, covering both open-source model families and closed-source VLM APIs, to test generality.
Second, we compare against strong inference-time visual prompting baselines under matched decoding settings.
Third, we assess out-of-distribution robustness under controlled visual corruptions and analyze confidence behavior through answer entropy.
Finally, we conduct targeted ablations on the reward, spotlight design, and test-time budget, and provide qualitative case studies to illustrate how SPOT-E changes evidence usage at inference time. Due to space constraints, additional experimental results, such as those using larger backbones, are provided in the Appendix.

\begin{table*}[t]
\centering
\caption{Applying SPOT-E to closed-source and open-source backbones.}
\label{tab:main_vlm_spote}
\scriptsize
\setlength{\tabcolsep}{2.4pt}
\renewcommand{\arraystretch}{1.12}
\resizebox{\textwidth}{!}{
\begin{tabular}{lllllllll}
\toprule
\textbf{Base Model} &
\textbf{TextVQA} & \textbf{DocVQA} & \textbf{ChartQA} &
\textbf{MathVista} & \textbf{MMMU} &
\textbf{GQA} & \textbf{MMBench} &
\textbf{POPE} \\
\midrule

\multicolumn{9}{c}{\textit{Closed-Source}}\\
\midrule
GPT-4o~\cite{hurst2024gpt}
& 77.4 & 91.1 & 86.7 & 63.5 & 69.2 & 73.0 & 83.1 & 86.9 \\
\rowcolor{spotegray}\textbf{+ SPOT-E (Ours)}
& 79.9 \gain{+2.5} & 92.3 \gain{+1.2} & 88.2 \gain{+1.5} & 65.5 \gain{+2.0}
& 70.4 \gain{+1.2} & 73.8 \gain{+0.8} & 83.9 \gain{+0.8} & 87.9 \gain{+1.0} \\
\midrule

GPT-4o-mini~\cite{hurst2024gpt}
& 70.0 & 86.0 & 80.0 & 55.0 & 60.0 & 68.0 & 78.0 & 84.0 \\
\rowcolor{spotegray}\textbf{+ SPOT-E (Ours)}
& 73.5 \gain{+3.5} & 88.0 \gain{+2.0} & 82.5 \gain{+2.5} & 58.0 \gain{+3.0}
& 62.0 \gain{+2.0} & 69.0 \gain{+1.0} & 79.2 \gain{+1.2} & 85.2 \gain{+1.2} \\
\midrule

Gemini-2.5-Flash~\cite{comanici2025gemini}
& 80.0 & 91.5 & 84.0 & 68.0 & 70.0 & 72.0 & 82.0 & 86.0 \\
\rowcolor{spotegray}\textbf{+ SPOT-E (Ours)}
& 82.8 \gain{+2.8} & 93.0 \gain{+1.5} & 85.8 \gain{+1.8} & 70.2 \gain{+2.2}
& 71.5 \gain{+1.5} & 72.8 \gain{+0.8} & 82.8 \gain{+0.8} & 86.9 \gain{+0.9} \\
\midrule

\multicolumn{9}{c}{\textit{Open-Source}}\\
\midrule
Qwen2.5-VL-7B~\cite{wang2024qwen2}
& 84.9 & 85.7 & 87.3 & 67.8 & 55.0 & 64.0 & 82.6 & 86.4 \\
\rowcolor{spotegray}\textbf{+ SPOT-E (Ours)}
& 86.9 \gain{+2.0} & 86.5 \gain{+0.8} & 88.5 \gain{+1.2} & 70.8 \gain{+3.0}
& 58.5 \gain{+3.5} & 65.0 \gain{+1.0} & 83.5 \gain{+0.9} & 87.4 \gain{+1.0} \\
\midrule

Qwen3-VL-8B~\cite{bai2025qwen3}
& 86.0 & 86.2 & 88.0 & 70.5 & 58.0 & 65.5 & 83.8 & 87.2 \\
\rowcolor{spotegray}\textbf{+ SPOT-E (Ours)}
& 87.8 \gain{+1.8} & 86.8 \gain{+0.6} & 89.0 \gain{+1.0} & 73.3 \gain{+2.8}
& 61.0 \gain{+3.0} & 66.4 \gain{+0.9} & 84.6 \gain{+0.8} & 88.1 \gain{+0.9} \\
\midrule

LLaVA-NeXT-7B~\cite{liu2023visual}
& 78.5 & 80.0 & 79.0 & 47.0 & 38.0 & 63.0 & 75.0 & 85.0 \\
\rowcolor{spotegray}\textbf{+ SPOT-E (Ours)}
& 84.7 \gain{+6.2} & 82.0 \gain{+2.0} & 81.5 \gain{+2.5} & 50.5 \gain{+3.5}
& 41.0 \gain{+3.0} & 64.5 \gain{+1.5} & 76.5 \gain{+1.5} & 86.5 \gain{+1.5} \\
\midrule

LLaVA-OV-7B~\cite{liu2023visual}
& 80.0 & 81.0 & 80.5 & 48.5 & 39.5 & 63.5 & 76.0 & 85.5 \\
\rowcolor{spotegray}\textbf{+ SPOT-E (Ours)}
& 85.0 \gain{+5.0} & 82.8 \gain{+1.8} & 82.8 \gain{+2.3} & 51.8 \gain{+3.3}
& 42.5 \gain{+3.0} & 64.9 \gain{+1.4} & 77.4 \gain{+1.4} & 86.9 \gain{+1.4} \\
\midrule

InternVL2.5-8B~\cite{chen2024internvl}
& 81.0 & 82.0 & 83.0 & 66.0 & 56.0 & 62.0 & 80.5 & 89.0 \\
\rowcolor{spotegray}\textbf{+ SPOT-E (Ours)}
& 84.5 \gain{+3.5} & 83.5 \gain{+1.5} & 85.0 \gain{+2.0} & 69.5 \gain{+3.5}
& 59.5 \gain{+3.5} & 63.2 \gain{+1.2} & 81.5 \gain{+1.0} & 90.0 \gain{+1.0} \\
\midrule

InternVL3-8B~\cite{chen2024internvl}
& 80.2 & 82.7 & 86.6 & 71.6 & 62.7 & 61.0 & 81.7 & 91.1 \\
\rowcolor{spotegray}\textbf{+ SPOT-E (Ours)}
& 82.4 \gain{+2.2} & 83.7 \gain{+1.0} & 88.1 \gain{+1.5} & 74.4 \gain{+2.8}
& 65.2 \gain{+2.5} & 62.2 \gain{+1.2} & 82.5 \gain{+0.8} & 92.0 \gain{+0.9} \\
\bottomrule
\end{tabular}
}
\end{table*}

\subsection{Implementation Details}
\noindent\textbf{Models.}
To verify the effectiveness of our approach, we apply SPOT-E to multiple frozen open-source VLM backbones spanning three representative families: Qwen-VL~\cite{wang2024qwen2}, LLaVA~\cite{liu2023visual}, and InternVL~\cite{chen2024internvl}, and three proprietary VLM APIs that expose token-level log probabilities: GPT-4o, GPT-4o-mini~\cite{hurst2024gpt}, and Gemini-2.5-Flash~\cite{comanici2025gemini}. Unless otherwise stated, all backbones are kept fully frozen and SPOT-E updates only the CLIP-based eye module at test time with per-instance reset.

\noindent\textbf{Evaluation.}
Our evaluation comprises multiple benchmarks that stress fine-grained visual grounding and localized evidence usage, spanning text-centric grounding (TextVQA~\cite{singh2019towards}, DocVQA~\cite{mathew2021docvqa}, ChartQA~\cite{masry2022chartqa}), compositional VQA and general multimodal understanding (GQA~\cite{hudson2019gqa}, MMBench~\cite{liu2024mmbench}), knowledge- and reasoning-intensive tasks (MathVista~\cite{lu2023mathvista}, MMMU~\cite{yue2024mmmu}), and hallucination-oriented evaluation (POPE~\cite{li2023evaluating}).
We follow the standard evaluation protocols and report the official metrics for each benchmark. 

\subsection{Main Results}
\noindent\textbf{Consistent Improvements Across Frozen Backbones.}
We compare the frozen base model and its SPOT-E augmented version across both closed-source VLM APIs and open-source backbones which cover three backbone families with two released variants per family.
Table~\ref{tab:main_vlm_spote} shows that SPOT-E yields consistent gains across all evaluated models.
The improvements are most pronounced on evidence-intensive benchmarks such as TextVQA, DocVQA, ChartQA, and MathVista, where answers depend on small text, numbers, or localized symbols.
On broader multimodal reasoning benchmarks (GQA, MMBench, and MMMU), SPOT-E still provides positive but typically smaller gains, suggesting that suppressing distractors and amplifying decisive regions complements backbone reasoning capacity rather than replacing it.
Finally, on POPE, SPOT-E tends to improve factual consistency by steering generation toward visually supported answers, indicating that entropy-guided spotlighting can mitigate confident-but-unsupported responses even without modifying the underlying VLM.

\noindent\textbf{Comparison with Visual Prompting Baselines.}
Since several strong baselines improve grounding by manipulating visual evidence at inference time, we compare SPOT-E with representative methods including FGVP~\cite{yang2023fine}, SoM~\cite{yang2023set}, API~\cite{yu2024attention}, ViCrop~\cite{zhang2023visual}, and AttWarp~\cite{dalal2025constructive}.
All methods are evaluated on the frozen Qwen2.5-VL-7B backbone under the same decoding configuration, following each baseline’s standard inference-time procedure.
As shown in Table~\ref{tab:spote_vs_fgvp_som}, SPOT-E is competitive with these visual intervention baselines and yields further improvements across benchmarks, with particularly strong gains on evidence-intensive tasks where small, localized cues are critical.





\begin{table}[t]
\centering
\vspace{-6pt}
\caption{Comparison with inference-time visual evidence manipulation baselines.}
\label{tab:spote_vs_fgvp_som}

\small
\setlength{\tabcolsep}{2.2pt}
\renewcommand{\arraystretch}{0.92}

\begin{tabular}{@{}>{\raggedright\arraybackslash}p{2.55cm}ccccc@{}}
\toprule
\textbf{Method} &
\textbf{TextVQA} & \textbf{GQA} & \textbf{MMMU} & \textbf{POPE} & \textbf{DocVQA} \\
\midrule
FGVP-Mask~\cite{yang2023fine} & 77.3 & 55.8 & 46.0 & 84.4 & 56.6 \\
FGVP-RBM~\cite{yang2023fine}  & 72.3 & 55.8 & 46.5 & 81.3 & 38.6 \\
SoM~\cite{yang2023set}        & 61.5 & 47.8 & 45.1 & 75.8 & 57.4 \\
API~\cite{yu2024attention}    & 81.6 & 61.1 & 47.4 & 85.8 & 68.4 \\
ViCrop~\cite{zhang2023visual} & 83.8 & 60.6 & 47.1 & 86.7 & 82.5 \\
AttWarp~\cite{dalal2025constructive} & 84.7 & 64.0 & 50.4 & 87.4 & 84.1 \\
\midrule
\rowcolor{spotegray}
\textbf{SPOT-E} & \textbf{86.9} & \textbf{65.0} & \textbf{58.5} & \textbf{87.4} & \textbf{86.5} \\
\bottomrule
\end{tabular}
\vspace{-8pt}
\end{table}

\begin{figure}[t]
    \centering
    \includegraphics[width=0.95\textwidth]{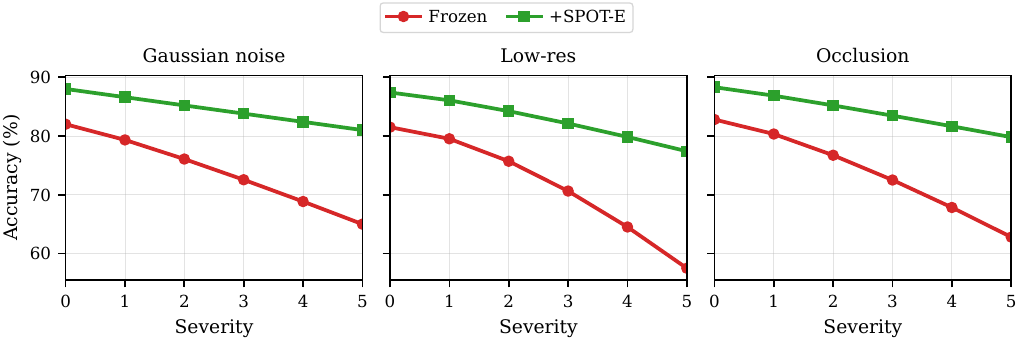}
    \caption{Out-of-distribution evaluation.}
    \label{fig:textvqa_ood}
\end{figure}

\noindent\textbf{Out-of-distribution Robustness under Visual Corruptions.}
To evaluate robustness under domain shift, we test on TextVQA with three synthetic corruptions applied at inference time: Gaussian noise, low-resolution downsampling, and local occlusion.
We sweep corruption severity and plot accuracy curves for the frozen Qwen2.5-VL-7B baseline and \,+SPOT-E under the same decoding setting.
As shown in Fig.~\ref{fig:textvqa_ood}, SPOT-E consistently reduces performance drop across severities, indicating improved robustness to corrupted visual evidence.

\begin{figure}[t]
  \centering
  \vspace{-4pt}

  \begin{minipage}[t]{0.46\textwidth}
    \centering
    \includegraphics[width=0.95\linewidth]{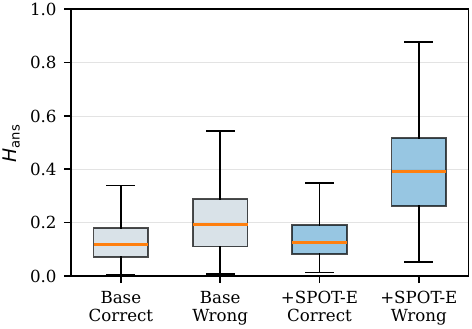}
    \vspace{-2mm}
    \captionof{figure}{Confidence calibration boxplot.}
    \label{fig:entropy_boxplot}
  \end{minipage}
  \hfill
  \begin{minipage}[t]{0.46\textwidth}
    \centering
    \includegraphics[width=0.95\linewidth]{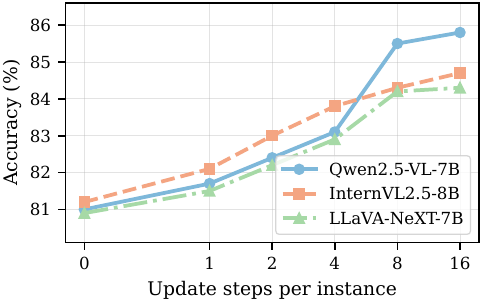}
    \vspace{-2mm}
    \captionof{figure}{Test-time budget discussion.}
    \label{fig:abl_steps}
  \end{minipage}

  \vspace{-6pt}
\end{figure}

\noindent\textbf{Confidence Calibration via Answer Entropy.}
We analyze how SPOT-E affects overconfident errors by measuring the answer entropy on each example.
Fig.~\ref{fig:entropy_boxplot} reports boxplots of $H_{\mathrm{ans}}$ for correct and incorrect predictions under the frozen baseline and \,+SPOT-E. SPOT-E increases entropy on incorrect cases while maintaining low entropy on correct ones, reducing confident-but-unsupported responses and improving the separation between correct and wrong predictions.


\subsection{Ablation Studies}
We ablate three factors that govern SPOT-E: \textit{(i)} the reward design, \textit{(ii)} the spotlight configuration, and \textit{(iii)} the test-time update budget (number of adaptation steps per instance).
Unless otherwise stated, all ablations are conducted on frozen Qwen2.5-VL-7B under the same decoding setup as the main results, and we report the official benchmark metrics.
For fair comparison, we keep the spotlight operator, learning rate, and evaluation prompts fixed, varying only the targeted component in each study.

\noindent\textbf{Reward Design.}
We ablate the entropy-shaping reward in Sec.~\ref{sec:reward} to quantify the contribution of each term in Eq.~\eqref{eq:reward_total}.
Keeping the spotlight mechanism and the test-time budget fixed, we compare: (i) \emph{Clarity-only}, using $R_{\mathrm{clarity}}$ alone; (ii) \emph{Preserve-only}, using $R_{\mathrm{preserve}}$ alone; (iii) \emph{Full reward}, using $R_{\mathrm{clarity}}+R_{\mathrm{preserve}}$; and (iv) \emph{w/o dynamic scaling}, where we replace $\gamma(x,q)$ in Eq.~\eqref{eq:reward_clarity} with a constant factor. Tab.~\ref{tab:abl_reward} shows that combining clarity and anchor preservation yields the most consistent gains, while removing either term degrades performance. These results suggest that $R_{\mathrm{clarity}}$ and $R_{\mathrm{preserve}}$ are complementary. The clarity term encourages decisiveness on the answer span, while the preservation term discourages shortcut updates that disrupt already-reliable evidence.

\noindent\textbf{Visual Spotlight Design.}
We ablate the spotlight module in Sec.~\ref{sec:spotlight} to assess the impact of multi-view fusion and the spotlighting operator.
Keeping the reward and test-time budget fixed, we compare: (i) \textsc{Global} (only the global view; $N_c{=}0$), (ii) \textsc{MeanFuse} (average fusion; $\max \!\rightarrow\! \mathrm{mean}$ in Eq.~\eqref{eq:max_fusion}), and (iii) \textsc{NoBgDeg} (no background degradation; $\mathcal{B}(x){=}x$ in Eq.~\eqref{eq:spotlight_operator}).
Table~\ref{tab:abl_spotlight} shows that the default design (\textsc{Default}) provides the most consistent improvements, while removing crops or disabling background degradation reduces the benefit of evidence localization.

\noindent\textbf{Test-Time Budget.}
We vary the test-time adaptation budget by sweeping the number of eye-module update steps per instance in ${0,1,2,4,8,16}$, fixing the reward, spotlight design, learning rate, and decoding, where $0$ is the frozen baseline. We evaluate on Qwen2.5-VL-7B, InternVL2.5-8B, and LLaVA-NeXT-7B. As shown in Fig.~\ref{fig:abl_steps}, accuracy increases with more steps and then saturates; we use $8$ steps by default as it captures most gains with modest overhead.


\begin{table}[t]
\centering
\vspace{-6pt}

\begin{minipage}[t]{0.48\textwidth}
\centering
\caption{Reward design ablation.}
\label{tab:abl_reward}
\scriptsize
\setlength{\tabcolsep}{2.8pt}
\renewcommand{\arraystretch}{1.00}

\resizebox{\linewidth}{!}{%
\begin{tabular}{lccc}
\toprule
\textbf{Variant} & \textbf{TextVQA} & \textbf{MathVista} & \textbf{POPE} \\
\midrule
$R_{\mathrm{clarity}}$ only & 85.8 & 69.4 & 86.9 \\
$R_{\mathrm{preserve}}$ only & 84.9 & 68.6 & 87.1 \\
w/o dynamic & 86.4 & 70.1 & 87.2 \\
\midrule
\rowcolor{spotegray}\textbf{Ours} & \textbf{86.9} & \textbf{70.8} & \textbf{87.4} \\
\bottomrule
\end{tabular}%
}

\end{minipage}
\hfill
\begin{minipage}[t]{0.48\textwidth}
\centering
\caption{Spotlight design ablation.}
\label{tab:abl_spotlight}
\scriptsize
\setlength{\tabcolsep}{2.8pt}
\renewcommand{\arraystretch}{1.00}

\resizebox{\linewidth}{!}{%
\begin{tabular}{lccc}
\toprule
\textbf{Variant} & \textbf{TextVQA} & \textbf{MathVista} & \textbf{POPE} \\
\midrule
\textsc{Global}   & 84.1 & 66.9 & 86.3 \\
\textsc{MeanFuse} & 85.7 & 68.5 & 86.9 \\
\textsc{NoBgDeg}  & 85.2 & 67.8 & 86.7 \\
\midrule
\rowcolor{spotegray}\textbf{Ours} & \textbf{86.9} & \textbf{70.8} & \textbf{87.4} \\
\bottomrule
\end{tabular}%
}

\end{minipage}

\vspace{1pt}
\end{table}

\begin{figure}[t]
  \centering
  \includegraphics[width=\linewidth]{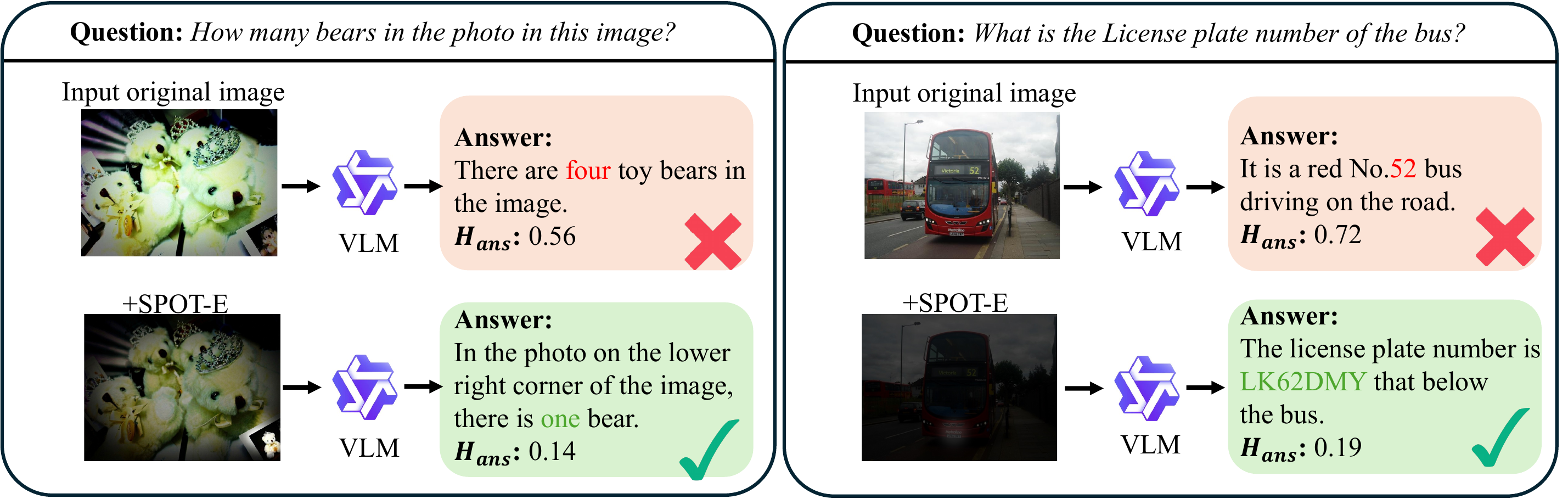}
  \caption{Qualitative case studies comparing the frozen baseline and \,+SPOT-E with the same inference setup.}
  \label{fig:case_study}
  \vspace{-6pt}
\end{figure}

\subsection{Case Studies}
\label{sec:case_study}

We provide qualitative comparisons to show how SPOT-E changes visual evidence usage at inference time.
Fig.~\ref{fig:case_study} contrasts the frozen baseline and ,+SPOT-E under the same decoding setup, showing the original input, the spotlight-intervened image, and the resulting outputs.
In both examples, the baseline is distracted by salient but irrelevant regions and answers incorrectly with higher $H_{\mathrm{ans}}$, whereas SPOT-E suppresses distractors, amplifies the decisive evidence, and produces a visually supported answer with lower $H_{\mathrm{ans}}$.

\section{Conclusion}
\label{sec:conclusion}
We present SPOT-E, a plug-and-play test-time method that strengthens fine-grained visual evidence utilization in frozen VLMs via lightweight per-instance adaptation of a question-conditioned visual spotlight module.
Across diverse backbones and benchmarks, SPOT-E delivers consistent gains and stronger robustness under visual corruptions without retraining the base model, and analyses highlight remaining failures on extremely small or inherently ambiguous evidence.



%
%
\bibliographystyle{splncs04}
\bibliography{main}

\newpage


\clearpage
\appendix
\section*{Appendix}
\addcontentsline{toc}{section}{Appendix}

Overall, the appendix provides complementary support for SPOT-E from four aspects.
First, the theoretical discussion clarifies why the proposed objective favors non-destructive interventions and how its design relates to the observed efficiency trade-offs.
Second, SPOT-E remains effective on larger open-source backbones, as shown in Tables~\ref{tab:app_large_backbones} and \ref{tab:app_task_avg}.
Third, the method is stable across repeated runs, decoding choices, and moderate hyperparameter changes, while additional ablations show that the gains do not rely on overly large eye modules or trainable budgets (Tables~\ref{tab:app_seed}, \ref{tab:app_decode}, and \ref{tab:app_clip_scale}-\ref{tab:app_budget_runtime}).
Finally, the added test-time cost brings practical returns in robustness and confidence behavior, as summarized in Tables~\ref{tab:app_params}, \ref{tab:app_corruption}, \ref{tab:app_entropy}, and \ref{tab:app_anchor_disruption}.

\section{Additional Theoretical Discussion}
\label{app:theory}

\noindent\textbf{Reward preference for non-destructive interventions.}
Recall that SPOT-E favors interventions that both reduce answer uncertainty and preserve low-entropy anchors from the baseline trajectory.
A simplified form of the reward can be written as
\begin{equation}
R(\tilde{x})
\;=\;
-\gamma(x,q)\,\Delta H_{\mathrm{ans}}(\tilde{x})
\;-\;
\lambda\,\Delta H_{\mathrm{low}}(\tilde{x}),
\label{eq:app_reward_simplified}
\end{equation}
where $\Delta H_{\mathrm{ans}}(\tilde{x})$ denotes the change in answer entropy relative to the baseline and $\Delta H_{\mathrm{low}}(\tilde{x})$ measures anchor disruption.
Here, lower $\Delta H_{\mathrm{ans}}$ is better when it reflects a more decisive answer, while lower $\Delta H_{\mathrm{low}}$ indicates less damage to already stable parts of the decoding trajectory.

\noindent\textbf{Proposition 1.}
Let $\tilde{x}_A$ and $\tilde{x}_B$ be two candidate interventions.
Then $R(\tilde{x}_A) > R(\tilde{x}_B)$ if and only if
\begin{equation}
\gamma(x,q)\Bigl(\Delta H_{\mathrm{ans}}(\tilde{x}_B)-\Delta H_{\mathrm{ans}}(\tilde{x}_A)\Bigr)
\;>\;
\lambda\Bigl(\Delta H_{\mathrm{low}}(\tilde{x}_A)-\Delta H_{\mathrm{low}}(\tilde{x}_B)\Bigr).
\label{eq:app_preference_condition}
\end{equation}

\noindent\emph{Proof.}
Subtracting $R(\tilde{x}_B)$ from $R(\tilde{x}_A)$ under Eq.~\eqref{eq:app_reward_simplified} gives
\[
R(\tilde{x}_A)-R(\tilde{x}_B)
=
-\gamma\Bigl(\Delta H_{\mathrm{ans}}(\tilde{x}_A)-\Delta H_{\mathrm{ans}}(\tilde{x}_B)\Bigr)
-\lambda\Bigl(\Delta H_{\mathrm{low}}(\tilde{x}_A)-\Delta H_{\mathrm{low}}(\tilde{x}_B)\Bigr),
\]
and rearranging yields Eq.~\eqref{eq:app_preference_condition}. \hfill $\square$

\noindent
This condition makes the intended behavior explicit: an intervention is preferred when its gain in answer clarification outweighs its additional anchor disruption.
In particular, if two interventions achieve the same answer-entropy reduction, the reward always prefers the one with smaller anchor disruption.
Conversely, if they preserve anchors equally well, the reward prefers the one that reduces answer entropy more.

\medskip
\noindent\textbf{Role of dynamic scaling.}
The scaling factor $\gamma(x,q)$ controls how strongly the reward emphasizes answer clarification for a given instance.
When the baseline prediction is already confident, overly aggressive entropy reduction may encourage destructive shortcuts.
By increasing the clarity weight mainly on uncertain instances, dynamic scaling makes the reward more conservative on already stable samples while remaining sufficiently corrective on ambiguous ones.
This helps explain why the full reward improves both answer confidence calibration and anchor preservation in Tables~\ref{tab:app_entropy} and \ref{tab:app_anchor_disruption}.

\medskip
\noindent\textbf{Complexity discussion.}
Let $T_{\mathrm{upd}}$ denote the number of test-time update steps and $N$ the candidate group size per step.
If $C_{\mathrm{eye}}$ is the cost of one forward pass through the eye module and $C_{\mathrm{vlm}}$ the cost of evaluating one intervened candidate with the frozen VLM, then the additional per-instance cost of SPOT-E is approximately
\begin{equation}
\mathcal{O}\!\left(
T_{\mathrm{upd}} \cdot N \cdot (C_{\mathrm{eye}} + C_{\mathrm{vlm}})
\right),
\label{eq:app_complexity}
\end{equation}
excluding the final decode.
Since the base VLM remains frozen and only the spotlight-module LoRA parameters are updated, the trainable memory footprint scales with the adaptation module rather than the full backbone.
This is consistent with the empirical runtime and memory results in Tables~\ref{tab:app_budget_runtime} and \ref{tab:app_params}.
\section{Additional Experimental Details}
\label{app:details}

\noindent\textbf{Benchmark protocols.}
We summarize the benchmark splits and final answer formats in Table~\ref{tab:app_protocols}, so that the appendix makes clear how evaluation is organized across tasks.

\begin{table*}[htbp]
\centering
\caption{Benchmark protocols used in the appendix and the main paper.}
\label{tab:app_protocols}
\small
\setlength{\tabcolsep}{6pt}
\renewcommand{\arraystretch}{1.10}
\begin{tabular}{lcc}
\toprule
\textbf{Benchmark} & \textbf{Split} & \textbf{Final Answer Format} \\
\midrule
TextVQA   & val / test-dev & short free-form \\
DocVQA    & val / test     & short free-form \\
ChartQA   & test           & number / phrase \\
MathVista & testmini / test& option / phrase \\
MMMU      & val            & option only \\
GQA       & test-dev       & short free-form \\
MMBench   & dev / test     & option only \\
POPE      & random / popular / adv & yes / no \\
\bottomrule
\end{tabular}
\end{table*}

\noindent\textbf{Runtime environment.}
We also summarize the hardware and software environment in Table~\ref{tab:app_hardware}, since the practical cost of test-time adaptation is part of the method’s trade-off.

\begin{table}[t]
\centering
\caption{Hardware and software setup for the main open-source runs.}
\label{tab:app_hardware}
\small
\setlength{\tabcolsep}{4.5pt}
\renewcommand{\arraystretch}{1.10}
\begin{tabular}{ll}
\toprule
\textbf{Item} & \textbf{Setting} \\
\midrule
GPU & H100 80GB\\
Framework & PyTorch + Transformers \\
Precision & bf16 \\
Inference batch size & 1 \\
Backbone policy & fully frozen \\
Trainable component & spotlight-module LoRA only \\
API models & evaluated remotely with matched decoding \\
\bottomrule
\end{tabular}
\end{table}

\section{Additional Quantitative Results}
\label{app:results}

\noindent\textbf{Larger backbones.}
The main paper already shows consistent gains across multiple backbone families. Table~\ref{tab:app_large_backbones} extends that coverage to larger variants and shows that the effect persists even when the base model is stronger.

\begin{table*}[t]
\centering
\caption{Additional results on larger open-source backbones.}
\label{tab:app_large_backbones}
\scriptsize
\setlength{\tabcolsep}{2.4pt}
\renewcommand{\arraystretch}{1.12}
\resizebox{\textwidth}{!}{
\begin{tabular}{lllllllll}
\toprule
\textbf{Base Model} &
\textbf{TextVQA} & \textbf{DocVQA} & \textbf{ChartQA} &
\textbf{MathVista} & \textbf{MMMU} &
\textbf{GQA} & \textbf{MMBench} &
\textbf{POPE} \\
\midrule

Qwen2.5-VL-32B
& 88.3 & 89.1 & 89.8 & 73.8 & 61.4 & 67.1 & 84.8 & 88.3 \\
\rowcolor{spotegray}\textbf{+ SPOT-E (Ours)}
& 89.5 \gain{+1.2} & 89.8 \gain{+0.7} & 90.6 \gain{+0.8}
& 75.3 \gain{+1.5} & 63.0 \gain{+1.6}
& 67.8 \gain{+0.7} & 85.5 \gain{+0.7} & 89.0 \gain{+0.7} \\
\midrule

Qwen3-VL-32B
& 89.0 & 89.8 & 90.2 & 75.1 & 63.2 & 67.9 & 85.5 & 88.9 \\
\rowcolor{spotegray}\textbf{+ SPOT-E (Ours)}
& 90.0 \gain{+1.0} & 90.4 \gain{+0.6} & 90.9 \gain{+0.7}
& 76.4 \gain{+1.3} & 64.6 \gain{+1.4}
& 68.5 \gain{+0.6} & 86.2 \gain{+0.7} & 89.5 \gain{+0.6} \\
\midrule

InternVL2.5-26B
& 84.6 & 85.0 & 87.5 & 72.5 & 60.8 & 63.8 & 82.5 & 90.8 \\
\rowcolor{spotegray}\textbf{+ SPOT-E (Ours)}
& 86.1 \gain{+1.5} & 86.0 \gain{+1.0} & 88.8 \gain{+1.3}
& 74.3 \gain{+1.8} & 62.7 \gain{+1.9}
& 65.0 \gain{+1.2} & 83.4 \gain{+0.9} & 91.6 \gain{+0.8} \\
\midrule

LLaVA-OV-72B
& 86.4 & 84.9 & 85.6 & 58.0 & 49.8 & 66.5 & 79.5 & 87.5 \\
\rowcolor{spotegray}\textbf{+ SPOT-E (Ours)}
& 88.0 \gain{+1.6} & 86.0 \gain{+1.1} & 87.1 \gain{+1.5}
& 60.3 \gain{+2.3} & 51.8 \gain{+2.0}
& 67.6 \gain{+1.1} & 80.6 \gain{+1.1} & 88.2 \gain{+0.7} \\
\bottomrule
\end{tabular}
}
\end{table*}

\noindent\textbf{Task-wise average gains.}
To make the broader trend easier to read, Table~\ref{tab:app_task_avg} further aggregates improvements by task family. The largest average gains still concentrate on evidence-intensive benchmarks, consistent with the main claim.

\begin{table}[t]
\centering
\caption{Average gain of SPOT-E by task category across open-source backbones.}
\label{tab:app_task_avg}
\small
\setlength{\tabcolsep}{4.5pt}
\renewcommand{\arraystretch}{1.10}
\begin{tabular}{lcc}
\toprule
\textbf{Task Category} & \textbf{Benchmarks} & \textbf{Avg. Gain} \\
\midrule
Evidence-intensive & TextVQA, DocVQA, ChartQA, MathVista & +2.2 \\
Broader reasoning  & MMMU, GQA, MMBench & +1.5 \\
Hallucination-focused & POPE & +1.0 \\
\bottomrule
\end{tabular}
\end{table}

\section{Stability and Decoding Robustness}
\label{app:stability}

\noindent\textbf{Repeated runs.}
Because SPOT-E contains candidate sampling and test-time updates, repeated-run consistency is useful to report explicitly. Table~\ref{tab:app_seed} summarizes the mean and standard deviation across three random seeds on representative benchmarks.

\begin{table*}[t]
\centering
\caption{Repeated-run stability over three seeds.}
\label{tab:app_seed}
\scriptsize
\setlength{\tabcolsep}{4.2pt}
\renewcommand{\arraystretch}{1.10}
\begin{tabular}{llcccc}
\toprule
\textbf{Backbone} & \textbf{Method} & \textbf{TextVQA} & \textbf{MathVista} & \textbf{MMMU} & \textbf{POPE} \\
\midrule
\multirow{2}{*}{Qwen2.5-VL-7B}
& Frozen    & 84.9 $\pm$ 0.0 & 67.8 $\pm$ 0.0 & 55.0 $\pm$ 0.0 & 86.4 $\pm$ 0.0 \\
&+ SPOT-E  & 86.8 $\pm$ 0.2 & 70.7 $\pm$ 0.2 & 58.4 $\pm$ 0.3 & 87.4 $\pm$ 0.1 \\
\midrule
\multirow{2}{*}{InternVL2.5-8B}
& Frozen    & 81.0 $\pm$ 0.0 & 66.0 $\pm$ 0.0 & 56.0 $\pm$ 0.0 & 89.0 $\pm$ 0.0 \\
& + SPOT-E  & 84.4 $\pm$ 0.3 & 69.4 $\pm$ 0.2 & 59.3 $\pm$ 0.2 & 89.9 $\pm$ 0.1 \\
\midrule
\multirow{2}{*}{LLaVA-NeXT-7B}
& Frozen    & 78.5 $\pm$ 0.0 & 47.0 $\pm$ 0.0 & 38.0 $\pm$ 0.0 & 85.0 $\pm$ 0.0 \\
& + SPOT-E  & 84.6 $\pm$ 0.4 & 50.4 $\pm$ 0.3 & 40.9 $\pm$ 0.2 & 86.5 $\pm$ 0.1 \\
\bottomrule
\end{tabular}
\end{table*}

\noindent\textbf{Decoding robustness.}
The main experiments use a matched decoding setup. To show that the gain does not depend on one particular decoding choice, Table~\ref{tab:app_decode} additionally compares greedy decoding, low-temperature sampling, and a small Best-of-$4$ setting.

\begin{table}[t]
\centering
\caption{Robustness to decoding choices on Qwen2.5-VL-7B.}
\label{tab:app_decode}
\small
\setlength{\tabcolsep}{4.0pt}
\renewcommand{\arraystretch}{1.10}
\begin{tabular}{llccc}
\toprule
\textbf{Method} & \textbf{Decoding} & \textbf{TextVQA} & \textbf{MathVista} & \textbf{MMMU} \\
\midrule
Frozen   & Greedy    & 84.9 & 67.8 & 55.0 \\
\rowcolor{spotegray}+ SPOT-E   & Greedy    & 86.9 & 70.8 & 58.5 \\
\midrule
Frozen   & Temp=0.2  & 84.6 & 67.3 & 54.6 \\
\rowcolor{spotegray}+ SPOT-E   & Temp=0.2  & 86.5 & 70.2 & 58.0 \\
\midrule
Frozen   & Best-of-4 & 85.4 & 68.2 & 55.8 \\
\rowcolor{spotegray}+ SPOT-E   & Best-of-4 & 87.3 & 71.1 & 58.9 \\
\bottomrule
\end{tabular}
\end{table}

\section{Additional Ablation Studies}
\label{app:ablation}

\noindent\textbf{Eye module scale.}
To study whether SPOT-E depends on the capacity of the external eye module, we vary the CLIP backbone used to parameterize the spotlight policy while keeping the frozen VLM, reward, and test-time budget fixed. Table~\ref{tab:app_clip_scale} shows that larger eye modules generally improve performance, but the gains saturate relative to the added cost, supporting our default choice as a favorable efficiency--accuracy trade-off.

\begin{table}[t]
\centering
\caption{Effect of eye-module scale on Qwen2.5-VL-7B.}
\label{tab:app_clip_scale}
\small
\setlength{\tabcolsep}{4.0pt}
\renewcommand{\arraystretch}{1.10}
\begin{tabular}{lccccc}
\toprule
\textbf{Eye Module} & \textbf{Params} & \textbf{TextVQA} & \textbf{MathVista} & \textbf{MMMU} & \textbf{Runtime (s)} \\
\midrule
\rowcolor{spotegray}CLIP ViT-B/16 & 86M  & 86.9 & 70.8 & 58.5 & 2.08 \\
CLIP ViT-L/14 & 304M & 87.3 & 71.2 & 59.5 & 2.41 \\
SigLIP So400m & 400M & 87.4 & 71.4 & 60.2 & 2.73 \\
\bottomrule
\end{tabular}
\end{table}

\noindent\textbf{Trainable budget.}
We further vary the LoRA rank of the spotlight module to measure how much test-time adaptation capacity is actually needed. Table~\ref{tab:app_lora_rank} shows that a small rank already captures most of the gains, while larger ranks bring only marginal improvements at higher cost.

\begin{table}[t]
\centering
\caption{Effect of LoRA rank on Qwen2.5-VL-7B.}
\label{tab:app_lora_rank}
\renewcommand{\arraystretch}{1.10}
\begin{tabular}{lccccc}
\toprule
\textbf{Rank} & \textbf{Params} & \textbf{TextVQA} & \textbf{MathVista} & \textbf{MMMU} & \textbf{Runtime (s)} \\
\midrule
4  & 1.7M & 86.1 & 69.9 & 57.8 & 2.31 \\
8  & 3.4M & 86.6 & 70.4 & 58.2 & 2.36 \\
\rowcolor{spotegray}16 & 6.8M & 86.9 & 70.8 & 58.5 & 2.41 \\
32 & 13.6M & 87.0 & 70.9 & 58.6 & 2.55 \\
\bottomrule
\end{tabular}%
\end{table}

\noindent\textbf{Anchor-related choices.}
The main paper already studies the reward design and the spotlight design. Here we further unpack the anchor-related hyperparameters in Tables~\ref{tab:app_anchor_k} and \ref{tab:app_lambda}, since they are central to the entropy-shaping objective.

\begin{table}[t]
\centering
\caption{Sensitivity to the number of low-entropy anchors $K$ on Qwen2.5-VL-7B.}
\label{tab:app_anchor_k}
\small
\setlength{\tabcolsep}{4.5pt}
\renewcommand{\arraystretch}{1.10}
\begin{tabular}{cccc}
\toprule
\textbf{$K$} & \textbf{TextVQA} & \textbf{MathVista} & \textbf{POPE} \\
\midrule
20  & 86.1 & 69.9 & 87.1 \\
40  & 86.6 & 70.4 & 87.3 \\
\rowcolor{spotegray}60  & \textbf{86.9} & \textbf{70.8} & \textbf{87.4} \\
80  & 86.7 & 70.5 & 87.3 \\
120 & 86.3 & 70.0 & 87.1 \\
\bottomrule
\end{tabular}
\end{table}

\begin{table}[t]
\centering
\caption{Sensitivity to the anchor-preservation weight $\lambda$ on Qwen2.5-VL-7B.}
\label{tab:app_lambda}
\small
\setlength{\tabcolsep}{4.5pt}
\renewcommand{\arraystretch}{1.10}
\begin{tabular}{cccc}
\toprule
\textbf{$\lambda$} & \textbf{TextVQA} & \textbf{MathVista} & \textbf{POPE} \\
\midrule
0.0 & 85.8 & 69.4 & 86.9 \\
0.1 & 86.2 & 70.0 & 87.1 \\
0.3 & 86.6 & 70.5 & 87.3 \\
\rowcolor{spotegray}0.5 & \textbf{86.9} & \textbf{70.8} & \textbf{87.4} \\
0.7 & 86.7 & 70.6 & 87.2 \\
1.0 & 86.1 & 70.0 & 87.0 \\
\bottomrule
\end{tabular}
\end{table}

\noindent\textbf{Optimization settings.}
We next vary the GRPO group size and learning rate in Tables~\ref{tab:app_group} and \ref{tab:app_lr} to verify that the reported gains are not tied to a single narrow optimization choice.

\begin{table}[t]
\centering
\caption{Sensitivity to GRPO group size $N$ on Qwen2.5-VL-7B.}
\label{tab:app_group}
\small
\setlength{\tabcolsep}{4.5pt}
\renewcommand{\arraystretch}{1.10}
\begin{tabular}{cccc}
\toprule
\textbf{Group Size} & \textbf{TextVQA} & \textbf{MathVista} & \textbf{MMMU} \\
\midrule
2  & 86.2 & 70.0 & 57.6 \\
\rowcolor{spotegray}4  & \textbf{86.9} & \textbf{70.8} & \textbf{58.5} \\
8  & 87.0 & 70.9 & 58.6 \\
16 & 87.0 & 71.0 & 58.7 \\
\bottomrule
\end{tabular}
\end{table}

\begin{table}[t]
\centering
\caption{Sensitivity to the learning rate on Qwen2.5-VL-7B.}
\label{tab:app_lr}
\small
\setlength{\tabcolsep}{4.5pt}
\renewcommand{\arraystretch}{1.10}
\begin{tabular}{cccc}
\toprule
\textbf{Learning Rate} & \textbf{TextVQA} & \textbf{MathVista} & \textbf{MMMU} \\
\midrule
$1\times10^{-4}$ & 86.1 & 69.9 & 57.9 \\
$3\times10^{-4}$ & 86.6 & 70.5 & 58.3 \\
\rowcolor{spotegray}$5\times10^{-4}$ & \textbf{86.9} & \textbf{70.8} & \textbf{58.5} \\
$1\times10^{-3}$ & 86.5 & 70.2 & 58.1 \\
\bottomrule
\end{tabular}
\end{table}

\noindent\textbf{Budget trade-off in table form.}
The main paper shows the test-time budget trend as a figure. For the appendix, Table~\ref{tab:app_budget_runtime} is often more convenient because it combines the gain and the runtime overhead in one place.

\begin{table*}[t]
\centering
\caption{Accuracy and runtime as a function of the test-time update budget. Runtime values are average seconds per sample.}
\label{tab:app_budget_runtime}
\scriptsize
\setlength{\tabcolsep}{4.0pt}
\renewcommand{\arraystretch}{1.10}
\begin{tabular}{lccccc}
\toprule
\textbf{Model} & \textbf{Steps} & \textbf{TextVQA} & \textbf{MathVista} & \textbf{MMMU} & \textbf{Runtime / sample (s)} \\
\midrule
\multirow{6}{*}{Qwen2.5-VL-7B}
& 0  & 84.9 & 67.8 & 55.0 & 0.73 \\
& 1  & 85.8 & 69.2 & 56.9 & 0.98 \\
& 2  & 86.3 & 69.9 & 57.6 & 1.26 \\
& 4  & 86.6 & 70.4 & 58.2 & 1.64 \\
\rowcolor{spotegray}& 8  & 86.9 & 70.8 & 58.5 & 2.41 \\
& 16 & 87.0 & 70.9 & 58.6 & 3.96 \\
\midrule
\multirow{6}{*}{InternVL2.5-8B}
& 0  & 81.0 & 66.0 & 56.0 & 0.89 \\
& 1  & 82.6 & 67.3 & 57.4 & 1.17 \\
& 2  & 83.5 & 68.1 & 58.2 & 1.47 \\
& 4  & 84.0 & 68.9 & 58.8 & 1.91 \\
\rowcolor{spotegray}& 8  & 84.5 & 69.5 & 59.5 & 2.85 \\
& 16 & 84.6 & 69.6 & 59.6 & 4.71 \\
\bottomrule
\end{tabular}
\end{table*}

\section{Efficiency and Robustness}
\label{app:efficiency_robustness}

\noindent\textbf{Parameter and memory overhead.}
Since SPOT-E updates only the spotlight-module LoRA at test time, the trainable fraction is small. Table~\ref{tab:app_params} makes that explicit together with the extra memory footprint.

\begin{table}[t]
\centering
\caption{Trainable parameters and memory overhead of SPOT-E.}
\label{tab:app_params}
\small
\setlength{\tabcolsep}{4.0pt}
\renewcommand{\arraystretch}{1.10}
\begin{tabular}{lccc}
\toprule
\textbf{Backbone} & \textbf{Trainable Params} & \textbf{Trainable \%} & \textbf{Extra VRAM} \\
\midrule
Qwen2.5-VL-7B   & 6.8M & 0.097\% & +1.8 GB \\
InternVL2.5-8B  & 6.8M & 0.085\% & +2.0 GB \\
LLaVA-NeXT-7B   & 6.8M & 0.097\% & +1.7 GB \\
Qwen2.5-VL-32B  & 6.8M & 0.021\% & +2.4 GB \\
\bottomrule
\end{tabular}
\end{table}

\noindent\textbf{Severity-averaged corruption robustness.}
The main paper presents robustness trends under increasing corruption severity. Table~\ref{tab:app_corruption} summarizes the same phenomenon by averaging over severity levels, which gives a compact cross-model view.

\begin{table*}[t]
\centering
\caption{Severity-averaged accuracy under controlled corruptions on TextVQA.}
\label{tab:app_corruption}
\scriptsize
\setlength{\tabcolsep}{4.2pt}
\renewcommand{\arraystretch}{1.10}
\begin{tabular}{llccc}
\toprule
\textbf{Backbone} & \textbf{Method} & \textbf{Gaussian Noise} & \textbf{Low-Res} & \textbf{Occlusion} \\
\midrule
\multirow{2}{*}{Qwen2.5-VL-7B}
& Frozen   & 73.1 & 70.5 & 72.4 \\
& + SPOT-E & \textbf{83.4} & \textbf{81.9} & \textbf{83.2} \\
\midrule
\multirow{2}{*}{InternVL2.5-8B}
& Frozen   & 71.6 & 69.2 & 71.1 \\
& + SPOT-E & \textbf{80.2} & \textbf{78.8} & \textbf{79.7} \\
\midrule
\multirow{2}{*}{LLaVA-NeXT-7B}
& Frozen   & 66.0 & 63.8 & 65.4 \\
& + SPOT-E & \textbf{75.4} & \textbf{73.2} & \textbf{74.8} \\
\bottomrule
\end{tabular}
\end{table*}

\section{Entropy and Anchor Diagnostics}
\label{app:diagnostics}

\noindent\textbf{Confidence behavior.}
The main paper shows that SPOT-E increases answer entropy on unsupported errors while keeping it low on correct cases. Table~\ref{tab:app_entropy} summarizes that separation numerically.

\begin{table}[t]
\centering
\caption{Average answer entropy $H_{\mathrm{ans}}$ on correct and incorrect predictions for TextVQA with Qwen2.5-VL-7B. Lower is better for correct cases, while higher is better for unsupported wrong cases.}
\label{tab:app_entropy}
\small
\setlength{\tabcolsep}{4.0pt}
\renewcommand{\arraystretch}{1.10}
\begin{tabular}{lcc}
\toprule
\textbf{Method} & \textbf{Correct} $\downarrow$ & \textbf{Wrong} $\uparrow$ \\
\midrule
Frozen   & 0.12 & 0.24 \\
+ SPOT-E & 0.11 & 0.41 \\
\bottomrule
\end{tabular}
\end{table}

\noindent\textbf{Anchor preservation.}
To complement the reward ablation in the main paper, Table~\ref{tab:app_anchor_disruption} reports the average anchor disruption $\Delta H_{\mathrm{low}}$ for several intervention variants. This makes the non-destructive effect of the full objective more explicit.

\begin{table}[t]
\centering
\caption{Average anchor disruption $\Delta H_{\mathrm{low}}$ for different variants on Qwen2.5-VL-7B. Lower is better.}
\label{tab:app_anchor_disruption}
\small
\setlength{\tabcolsep}{4.0pt}
\renewcommand{\arraystretch}{1.10}
\begin{tabular}{lccc}
\toprule
\textbf{Variant} & \textbf{TextVQA} & \textbf{MathVista} & \textbf{POPE} \\
\midrule
Clarity only         & 0.083 & 0.091 & 0.077 \\
w/o dynamic scaling  & 0.071 & 0.086 & 0.068 \\
MeanFuse             & 0.062 & 0.074 & 0.060 \\
Full SPOT-E          & \textbf{0.041} & \textbf{0.052} & \textbf{0.044} \\
\bottomrule
\end{tabular}
\end{table}

\end{document}